\documentclass[sigconf,screen]{acmart}

\usepackage{multirow}

\AtBeginDocument{%
  \providecommand\BibTeX{{%
    \normalfont B\kern-0.5em{\scshape i\kern-0.25em b}\kern-0.8em\TeX}}}

\setcopyright{acmcopyright}
\copyrightyear{2022}
\acmYear{2022}
\acmDOI{xxx.xxx}

\acmConference[MM'22]{the 30th ACM International Conference on Multimedia, 2022, Lisbon, Portugal}
\acmPrice{15.00}
\acmISBN{978-1-4503-XXXX-X/18/06}
\renewcommand\footnotetextcopyrightpermission[1]{}
\usepackage[misc]{ifsym}
\begin{document}

\title{ShowFace: Coordinated Face Inpainting with Memory-Disentangled Refinement Networks}
\pagestyle{empty}

\author{
Zhuojie Wu\textsuperscript{1},
Xingqun Qi\textsuperscript{1}, 
Zijian Wang\textsuperscript{1},
Wanting Zhou\textsuperscript{1}, 
Kun Yuan\textsuperscript{2}, 
Muyi Sun\textsuperscript{3, $\textrm{\Letter}$}, 
Zhenan Sun\textsuperscript{3}
}
\affiliation{
\institution{
    \textsuperscript{1}School of AI, BUPT,
    \textsuperscript{2}Kuaishou Technology,
    \textsuperscript{3}CRIPAC, NLPR, CASIA
    }
\country{}
}
\email{{zhuojiewu, xingqunqi, wangzijianbupt, wanting.zhou}@bupt.edu.cn}
\email{yuankun03@kuaishou.com, muyi.sun@cripac.ia.ac.cn, 
znsun@nlpr.ia.ac.cn}


\renewcommand{\shortauthors}{}

\begin{abstract}
Face inpainting aims to complete the corrupted regions of the face images, which requires coordination between the completed areas and the non-corrupted areas.
Recently, memory-oriented methods illustrate great prospects in the generation related tasks by introducing an external memory module to improve image coordination.
However, such methods still have limitations in restoring the consistency and continuity for specific facial semantic parts.
In this paper, we propose the coarse-to-fine Memory-Disentangled Refinement Networks (MDRNets) for coordinated face inpainting, in which two collaborative modules are integrated, Disentangled Memory Module (DMM) and Mask-Region Enhanced Module (MREM). 
Specifically, the DMM establishes \textbf{a group of disentangled memory blocks to store the semantic-decoupled face representations},
which could provide the most relevant information to refine the semantic-level coordination.
The MREM involves a \textbf{masked correlation mining mechanism} to enhance the feature relationships into the corrupted regions, which could also make up for the correlation loss caused by memory disentanglement. 
Furthermore, to better improve the inter-coordination between the corrupted and non-corrupted regions and enhance the intra-coordination in corrupted regions, we design \textbf{InCo$^2$ Loss, a pair of similarity based losses} to constrain the feature consistency.
Eventually, extensive experiments conducted on CelebA-HQ and FFHQ datasets demonstrate the superiority of our MDRNets compared with previous State-Of-The-Art methods. 
\vspace{-2mm}
\end{abstract}

\vspace{-10mm}
\ccsdesc[500]{Computing methodologies~Computer vision\vspace{-2mm}}

\keywords{
\vspace{-4mm}
Face inpainting; Disentangled memory; Masked correlation mining
\renewcommand{\thefootnote}{}
\footnotetext{$\textrm{\Letter}$ Corresponding author.\vspace{-1mm}}
}

\maketitle

\vspace{2mm}
\begin{figure}[h]
  \centering
  \includegraphics[width=0.95\linewidth]{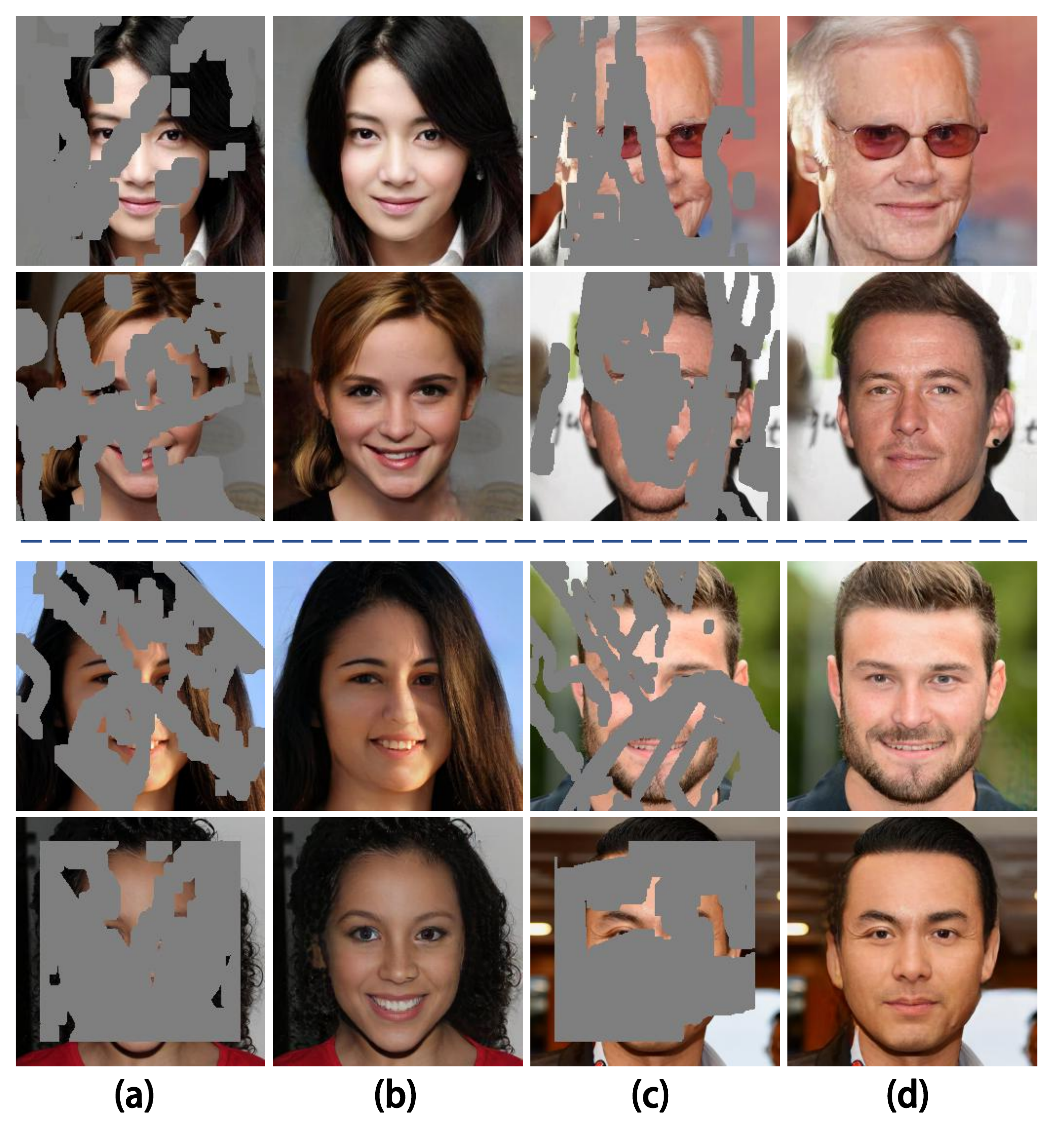}
  \vspace{-2mm}
  \caption{Randomly sampled results from our MDRNets which could achieve coordinated and realistic face inpainting with large masked regions. Column (a) and (c) are the network inputs. Column (b) and (d) are the corresponding face inpainting results through MDRNets. 
  The first two rows are from the CelebA-HQ\cite{karras2018progressive} dataset, and the last two rows are from FFHQ\cite{karras2019style} dataset. Zoom in for better details.}
  \Description{The MDRNets complete results with mask size of 50$\%$ to 60$\%$.}
  \label{figure1}
\end{figure}

\vspace{-2mm}
\section{Introduction}
Face inpainting is an ill-posed problem, which aims to restore the corrupted regions with coordinated contents as long as they appear plausible\cite{li2017generative,zhou2020learning,li2020learning,wang2022ft,yu2019free}. 
Since the corrupted regions traverse multiple semantic parts, the coordinated inpainting generally requires the consistency within each semantic and the coordination between different semantic parts.
Recently, face inpainting has shown great potential in sufficient real-world applications such as interactive face editing and occluded face recognition. 
However, it is still challenging for coordinated face recovery when the missing regions are large or the face contents are complex.
Therefore, face inpainting has been a continuous hot spot in the field of face generation.
Figure \ref{figure1} shows this inpainting task and illustrates several results of our method, which achieve coordinated face inpainting with large masked regions.

From the perspective of technology, face inpainting could be roughly divided into two categories in recent years: patch-based methods \cite{barnes2009patchmatch} and deep learning based methods \cite{li2017generative}. 
Specifically, the patch-based methods \cite{zhuang2009patch,tang2009face,yang2017high} generally sample image patches from the remaining image regions and fuse these patches to recover the missing areas, which could synthesize highly-textured complement contents. 
However, these low-level image patches will introduce inconsistency in corrupted regions.
Such methods will fail to generate semantically reasonable results due to the lack of high-level image understanding.
Meanwhile, the patch matching process is usually time consuming and laborious.
Furthermore, when the corrupted regions are large, the completed images will become smooth due to the insufficient remaining patches.


Recently, great progress has been made in image inpainting tasks with the remarkable development of Generative Adversarial Networks(GANs) \cite{goodfellow2014generative}.
Meanwhile, numerous researchers employ GANs in face inpainting\cite{yu2018generative,li2017generative,xie2019image,yang2020deep,liu2021deflocnet}.
For the specificity of the inpainting task, the vanilla convolution is upgraded in  \cite{liu2018image,yu2019free,xie2019image}, which change the conventional convolution mechanism and pay more attention to valid pixels.
Nevertheless, the above approaches expose a common drawback in recovering the image global structure.
Therefore, many studies improve the network to better recover the global structure by introducing relevant structural priors \cite{nazeri2019edgeconnect,peng2021generating,guo2021image,wan2021high,yu2021diverse}.
However, these low-level structural priors are difficult to obtain, under the large corrupted regions.
Meanwhile, the images synthesized by the above methods often lack detailed texture.
Thus, some researchers \cite{yu2018generative,liu2019coherent,xu2021texture} combine the advantages of the patch-based method and deep learning based method, which deliver the
inpainting contents with both detailed textures and plausible semantics. 
However, there is still a drawback of image coordination between the semantics in different patches.

Inspired by the above studies, we propose the coarse-to-fine Memory-Disentangled Refinement Networks (MDRNets) for coordinated face inpainting. 
The entire architecture of MDRNets can be summarized as \textbf{coarse network, disentangled memory module, masked correlation mining, and guided refinement network.} 
To begin with, the encoder-decoder based coarse network generates a coarse global face, which could produce reasonable global structural priors.
Then, we design a \textbf{Disentangled Memory Module (DMM)} to store the semantic-aware decoupled face latent vectors from the non-corrupted regions, in which a group of dynamic disentangled memory blocks are established. 
With this design, the disentangled semantic-level latent vectors ensure the coordination within each semantic.
Next, we propose a \textbf{Mask-Region Enhanced Module (MREM)} to enhance the feature relationships into the corrupted regions, which also makes up for the correlation loss caused by the memory disentanglement. 
The MREM involves a masked correlation mining mechanism to compute the relationships between the completed and the non-corrupted regions.
At last, in the guided refinement network, we  utilize the output of MREM to refine the coarse face through the SPADE \cite{park2019semantic} method.

Moreover, for improving the intra-coordination in the corrupted regions and the inter-coordination between corrupted and non-corrupted regions, we design \textbf{InCo$^2$ Loss, a pair of similarity based losses} to constrain the feature consistency.
Specifically, we construct two types of similarity matrices to mine deeper feature correlations from the corrupted and non-corrupted regions.

Overall, our method could \textbf{Show the Face}, no matter how many facial areas are corrupted. 
The main contributions can be summarized as follows:
\begin{itemize}
\item We propose Memory-Disentangled Refinement Networks (MDRNets) for coordinated face inpainting.
\item We design two collaborative modules, Disentangled Memory Module (DMM) and Mask-Region Enhanced Module (MREM), which achieve the memory disentanglement for semantic-level coordination and enhance the feature relationships for face inpainting.
\item We propose InCo$^2$ Loss, a pair of similarity based losses to further improve the inter-coordination between the corrupted and non-corrupted regions and the intra-coordination in corrupted regions. 
\item Both the qualitative and quantitative results on CelebA-HQ and FFHQ datasets demonstrate the effectiveness of the our method which achieves State-Of-The-Art performance.
\end{itemize}

\begin{figure*}[h]
  \centering
  \includegraphics[scale=0.38]{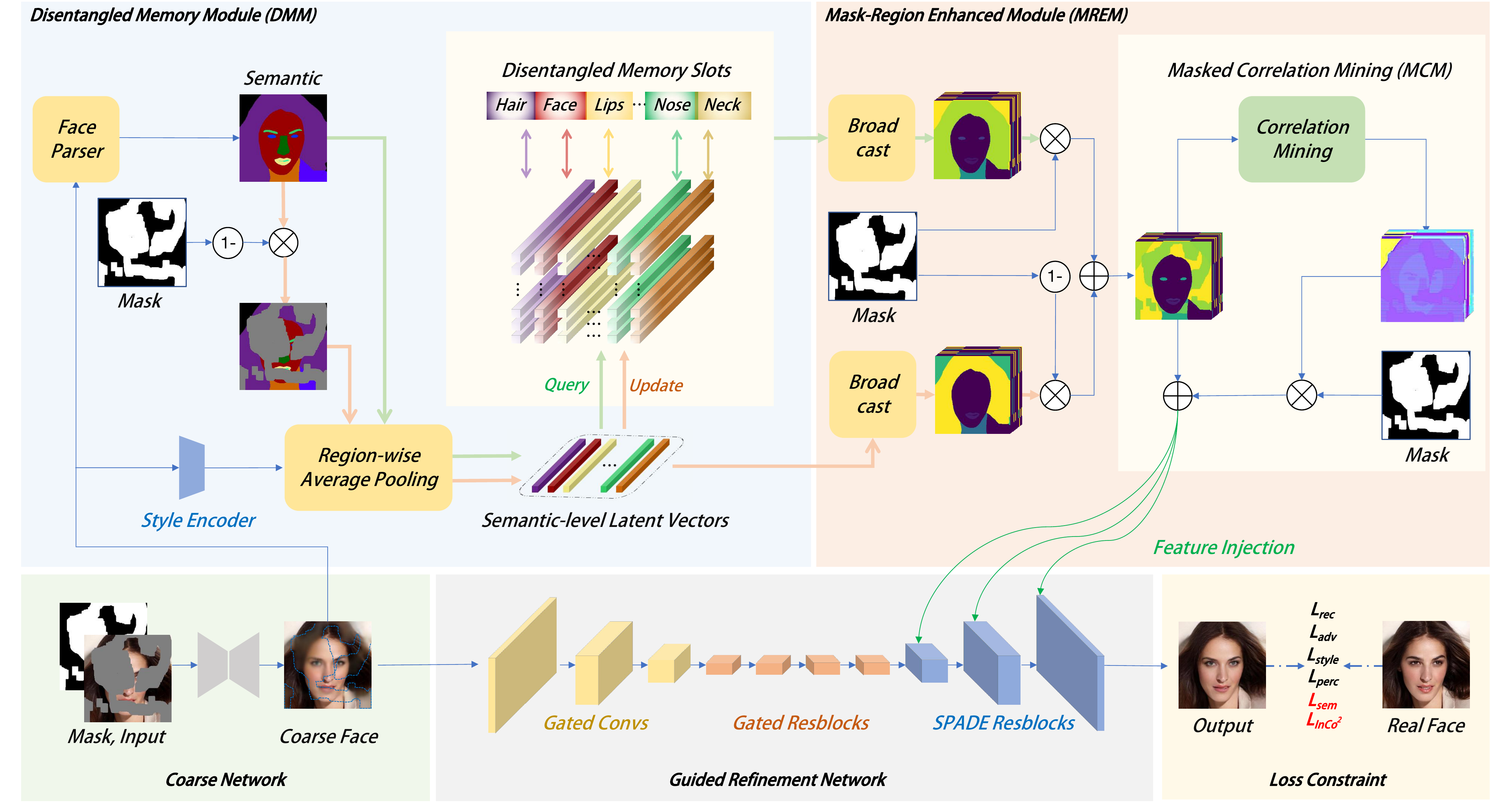}
  \caption{An overview of Memory-Disentangled Refinement Networks (MDRNets). 
  Given a masked image and the corresponding mask (black=0, white=1), we first employ the Coarse Network to generate the coarse global face. 
  Then, the DMM stores the semantic-level latent vectors from non-corrupted regions (PINK LINE in DMM)
  and provides the most relevant features according to the semantic-aware queries from the coarse face (GREEN LINE in DMM).
  Next, the MREM fuses the features from the DMM and constructs a correlation map to enhance the correlations into the corrupted regions. 
  Finally, the fused features after MREM are injected into the Guided Refinement Network to get the final result. 
  $\textcircled{1-}$, $\oplus$ and $\otimes$ denote the operations of 1-Mask, element-wise addition and element-wise multiplication respectively.}
  \Description{}
  \label{figure2}
\end{figure*}

\section{Related Work}
\subsection{Face Inpainting}
Face inpainting has made tremendous progress in the past few years. 
In previous patch-based methods, Zhuang \emph{et al.} \cite{zhuang2009patch} and Tang \emph{et al.} \cite{tang2009face} extract prototypical image patches to fill the missing areas. 
Xu \emph{et al.}\cite{xu2021texture} utilize the similarity matrix to seek patches for consistent texture generation. 
However, the patch-based methods are difficult to find suitable contents when the corrupted regions is foreground and large. 
Then, great efforts are made in early deep learning methods to maintain the image consistency and restore irregular masks.
Yu \emph{et al.} \cite{yu2018generative} establish the contextual relationship into the face inpainting networks.
Liu \emph{et al.} \cite{liu2018image} propose partial convolution for irregular mask to filter out invalid pixels.
Yu \emph{et al.} \cite{yu2019free} design a learnable dynamic feature selection mechanism, which generalizes the partial convolution. 
However, these early deep learning methods are limited in maintaining global consistency among face components, and the completed areas are generally blurry.
Recently, some methods are designed to integrate the face priors or new network architectures.
Li \emph{et al.} \cite{li2020learning} propose SymmFCNet, which use the symmetry of face to recover facial details.
Liu \emph{et al.} \cite{liu2021pd} introduce probabilistic diversity map, which controls the diversity extent of the completed faces.
Peng \emph{et al.} \cite{peng2021generating} and Guo \emph{et al.} utilize facial structure and texture constraints to guide the inpainting network.
Wan \emph{et al.} \cite{wan2021high} and Yu \emph{et al.} \cite{yu2021diverse} employ autoregressive transformers to inpaint diverse faces. 
However, these methods generally ignore the modeling of the facial internal correlations, and limit the refinement of the specific facial semantic regions.
In this paper, we propose Memory-Disentangled Refinement Networks (MDRNets) for coordinated face inpainting and enhance the feature correlations in the networks.

\subsection{Memory Networks}
Extensive deep learning methods possess the ability of memory, such as RNN \cite{mikolov2010recurrent}, LSTM\cite{hochreiter1997long} and GRU\cite{cho2014learning}. 
However, they are all limited in the long-term memory of information.
To overcome this shortcoming, Weston \emph{et al.}  \cite{Weston2015MemoryN} first propose memory networks, which employ explicit storage and attention mechanism to model the long-term information more effectively.
And due to the high efficiency for feature storage, memory networks have become popular in the field of image generation.
Yoo \emph{et al.} \cite{yoo2019coloring} present a memory-augmented colorization network to produce high-quality image colorization with limited data.
Huang \emph{et al.} \cite{huang2021memory} employ a dynamic memory block to record the prototypical patterns of rain degradations for rain removal.
Zhu \emph{et al.} \cite{zhu2019dm} introduce a multimodal memory module to refine blurred images for text-to-image generation.
Qi \emph{et al.} \cite{qi2021latent} design a latent memory unit to preserve the core storyline and history information for visual storytelling.
In the face inpainting task, Xu \emph{et al.} \cite{xu2021texture} firstly propose a patch-based texture memory to enhance the completed image texture.
However, these above methods either select static image patches to build the memory, or store the features into a unified memory block, which bring limitations in restoring the consistency and continuity for image semantic, especially in each specific semantic part.
In this paper, we introduce a group of dynamic disentangled memory blocks to store the semantic-decoupled face representations for coordinated face inpainting.

\subsection{Correlation Mining}

Correlation mining plays a significant role in computer vision \cite{yu2018generative,hou2021coordinate,tung2019similarity,liu2020structured,lee2020reference}.
Hou \emph{et al.} \cite{hou2021coordinate} establish a coordinate attention map to capture long-range dependencies along spacial direction.
Tung \emph{et al.} and Liu \emph{et al.} \cite{tung2019similarity,liu2020structured} construct correlation coefficients to constrain class-aware relationships, which improve the performance of the student network.
Lee \emph{et al.} \cite{lee2020reference} propose a dense correlation mapping, which transfers information from reference image to sketch for sketch colorization. 
Zhang \emph{et al.} \cite{zhang2020cross} introduce spatial correlation fields, which enable cross-domain image-to-image translation.
Liu \emph{et al.} \cite{liu2021adaattn} compute the correlation maps between the content and style features, which adaptively normalizes the content features to generate a natural output for style transfer. 
Mou \emph{et al.} \cite{mou2021dynamic} propose a dynamic attentive graph learning model, which establishes dynamic non-local correlations to balance over-smooth and over-sharp artifacts during image restoration.
Inspired by the correlation mining strategy, in this paper, we propose a masked correlation mining mechanism to enhance the feature correlations into the corrupted regions. What's more, a pair of similarity based losses are introduced to constrain the feature consistency.

\begin{figure*}[h]
  \centering
  \includegraphics[scale=0.6]{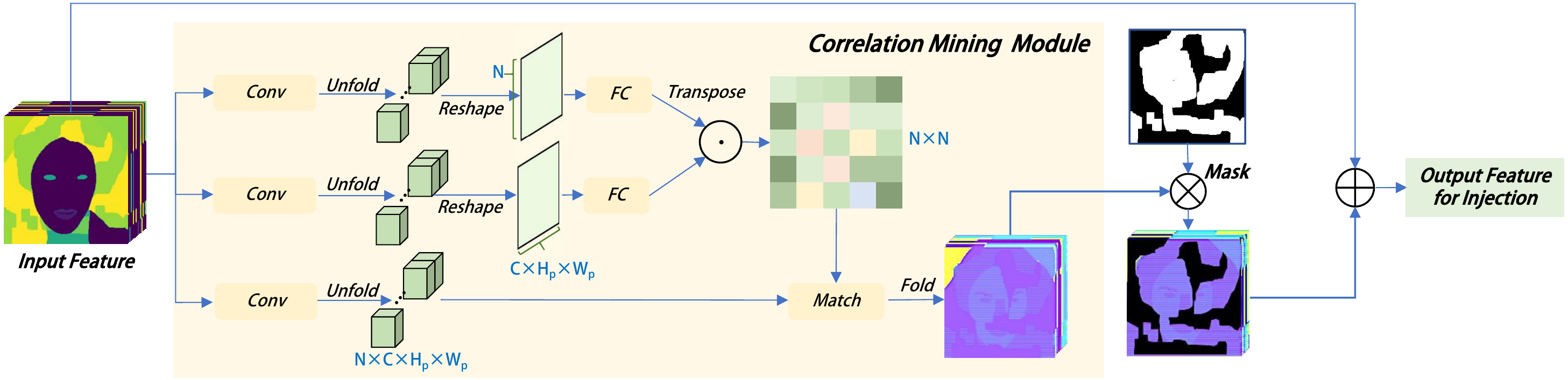}
  \caption{Detailed illustration of the Masked Correlation Mining (MCM), in which the correlation mining module computes the correlations within the features and the mask multiplication preserves the feature enhancement of the corrupted regions. 
  Finally, the correlation-enhanced corrupted regions are fused with the input features.
  $\odot$, $\oplus$ and $\otimes$ denote the dot product, element-wise addition and element-wise multiplication respectively.}
  \Description{}
  \label{figure3}
\end{figure*}

\section{METHOD}
In this section, we present our method in detail.
To begin with, we introduce the overall MDRNets. Then, we give the details of the specific components in the networks, especially the DMM and MREM. 
Finally, the total objective functions of this model are described. 

\subsection{Overview}

The proposed coarse-to-fine framework of MDRNets is shown in Figure \ref{figure2}.
To begin with, given a masked image $I$ and the corresponding mask $M$, we employ the pre-trained partial convolution based coarse network to generate the coarse global face $P$.
Then we leverage the face parser \cite{yu2018bisenet} to obtain the corresponding semantic map $S$ of the coarse result.
To finely recover each semantic part of the face and maintain semantic coordination, the DMM (i.e. the memory $\mathbb{M}$) is proposed to store the semantic-aware latent vectors $V$ from the non-corrupted regions, which are extracted by employing region-wise average pooling \cite{zhu2020sean} beyond the Style Encoder and the masked $S$. 
Then, the most relevant memory slots in $\mathbb{M}$ could be retrieved using the semantic-level latent vectors $Q$ of the coarse face as queries.
To enhance the feature relationships into the corrupted regions, the MREM is proposed to construct a correlation map, which could fuse the features of the non-corrupted regions into the corrupted regions.
Eventually, the generated features after MREM are injected into the Guided Refinement Network through SPADE \cite{park2019semantic} to get the coordinated face $\hat{I}$.
In the following, each component of our method is introduced in detail.

\subsection{Disentangled Memory Module}

To generate semantic-level coordinated faces, we propose the DMM to establish a group of disentangled memory blocks, which stores the semantic-decoupled face representations.
As illustrated in Figure \ref{figure2}, we employ the Style Encoder to extract the style feature maps $F_{s}\in \mathbb{R}^{c\times h\times w}$ from the coarse face $P$.
Meanwhile, the face parser is used to obtain the corresponding semantic map $S$ from $P$, which contains 14 different semantic categories in face images (\emph{e.g., skin, eye}). 
Then, we employ semantic-wise (region-wise) average pooling \cite{zhu2020sean} to obtain latent vectors $Q\in \mathbb{R}^{n\times c}$ and $V\in \mathbb{R}^{n\times c}$, where n represents the number of semantic categories.

\textbf{Memory blocks.}
We establish the disentangled memory blocks (14 blocks for 14 semantic categories) to store the semantic-aware $V$, which represents the non-corrupted and accurate latent representations of the facial parts.
Specifically, the proposed memory $\mathbb{M}\in \mathbb{R}^{n\times m \times c}$ consists of $n = 14$ memory blocks, in which each memory block contains $m$ memory slots $e_{ij}\in \mathbb{R}^{c}$. 
Taking each semantic-level latent vector in $Q$ as a query, we could retrieve a relevant representation from its corresponding memory block. 
This memory-based representation is obtained by integrating the $m$ semantic-related memory slots with soft scores.
Meanwhile, we could update the semantic-level $\mathbb{M}$ by the semantic-aware $V$.

\textbf{Memory Updating.}
The update of memory $\mathbb{M}$ is based on the similarity between the latent vectors in $Q$ and the corresponding memory slots. 
To begin with, we compute the $i$-th semantic cosine similarity $\gamma_{ij}$ between $Q_{i}$ and the $j$-th memory slot $e_{ij}$, defined as
\begin{equation}
\gamma_{ij}=\frac{e_{ij}Q_{i}^{T}}{\left\| e_{ij}\right\|\left\| Q_{i}\right\|} \label{eq1}
\end{equation}
Then, we retrieve the memory slot $e_{i \phi _{j}}$, which is most relevant with $Q_{i}$ in each training batch.
\begin{equation}
k_{j}=\underset{i}{argmax} (\gamma_{ij}) \label{eq2}
\end{equation}
To ensure the authenticity of the memory, we use the \textbf{non-corrupted latent vector $V_{i}$} from the non-corrupted regions in each training batch to update the memory slot $e_{i k_{j}}$, \textbf{which is also most similar and shares the same semantic with the query $Q_{i}$}.
\begin{equation}
e_{i k_{j}}\overset{}{\leftarrow} \alpha e_{i k_{j}} + (1-\alpha )V_{i}  \label{eq3}
\end{equation}
where $\alpha\in[0,1]$ is a decay rate. 

\textbf{Memory Reading.}
After updating the memory $\mathbb{M}$, we reconstruct memory-based latent vectors $\hat{Q}_{i}$, which is most relevant to $Q_{i}$. 
What's more, we employ soft scores to aggregate memory slots for end-to-end training. 
To begin with, the cosine similarity matrix $\Upsilon=\left\{\gamma_{ij}|i=1,...,n,j=1,...,m\right\}$ is computed by Eq.\ref{eq1} again. 
Then, the soft scores $A=\left\{a_{ij}|i=1,...,n,j=1,...,m\right\}$ are formulated by a softmax operation.
\begin{equation}
a_{ij}=\frac{exp(\gamma _{ij})}{\sum _{j=1}^{m}exp(\gamma _{ij})}  \label{eq4}
\end{equation}
Finally, the memory-based latent vectors $\hat{Q}_{i}$ is constructed by aggregating memory slots with the soft scores.
\begin{equation}
\hat{Q}_{i}=\sum_{j}^{m}a_{ij}e_{ij} \label{eq5}
\end{equation}

Furthermore, different from previous methods\cite{zhu2019dm,huang2021memory,xu2021texture},in each updating or reading process, our method employs local slots (semantic-aware representations) for similarity calculation, rather than global slots (representations), which can be more efficient.
To sum up, we propose a \textbf{Dynamic Disentangled Efficient} memory mechanism.

\subsection{Mask-Region Enhanced Module}
To enhance the feature representation of the corrupted regions, we propose MREM, which consists of feature fusion and Masked Correlation Mining (MCM). 
To begin with, we broadcast memory-based latent vectors $\hat{Q}$ to semantic map $S$, which obtain memory-based feature maps.
Meanwhile, we obtain $F_{V}$ by broadcasting $V$ to semantic map $S$.  
Then, we employ the mask to achieve feature fusion, which ensures the fused features both come from the ``real” image features and share great similarity at the semantic level. 
The above processes are shown in Figure \ref{figure2}.

To focus on feature relationships, we further design MCM, which consists of a correlation mining module and mask multiplication.
The correlation mining module contains three branches, in which the first two branches compute the correlations within the features and then match the third branch. 
To begin with, We apply the 1 $\times$ 1 convolutional layer to transform the input features into two independent representations, and then utilize the unfold operation to extract N feature patches $\mathcal{P}\in \mathbb{R}^{C\times H_{p}\times W_{p}}$. 
Next, each feature patch is reshaped into a feature vector. 
The similarity matrix $\Phi \in \mathbb{R}^{N\times N}$ representing the correlations between each patch can be computed by dot product. 
Thus, we could update each patch by the similarity matrix $\Phi$. 
Through the correlation mining, the feature relationships are enhanced at the image-level, which also makes up for the correlation loss caused by memory disentanglement.
Finally, the mask multiplication preserves the feature enhancement of the corrupted regions. 
Then, the correlation-enhanced corrupted regions are fused with the input features by element-wise addition.
After MREM, which mines the correlation within the features and enhances the feature representation of the corrupted regions, we obtain the coordinated feature maps for the final injection. 

\subsection{Guided Refinement Network}

The Guided Refinement Network consists of gated convolutional layers \cite{yu2019free}, gated ResBlocks, and SPADE ResBlocks \cite{park2019semantic}. 
At first, we encode the $P$ to provide the texture of the non-corrupted regions for the final face generation. 
Then, the fused features $F_{f}$ after MREM are injected into the Guided Refinement Network by SPADE as shown in Figure \ref{figure2}, which facilitate the final coordinated face. 

\subsection{Objective Functions}

\begin{figure}[h]
  \centering
  \includegraphics[width=1\linewidth]{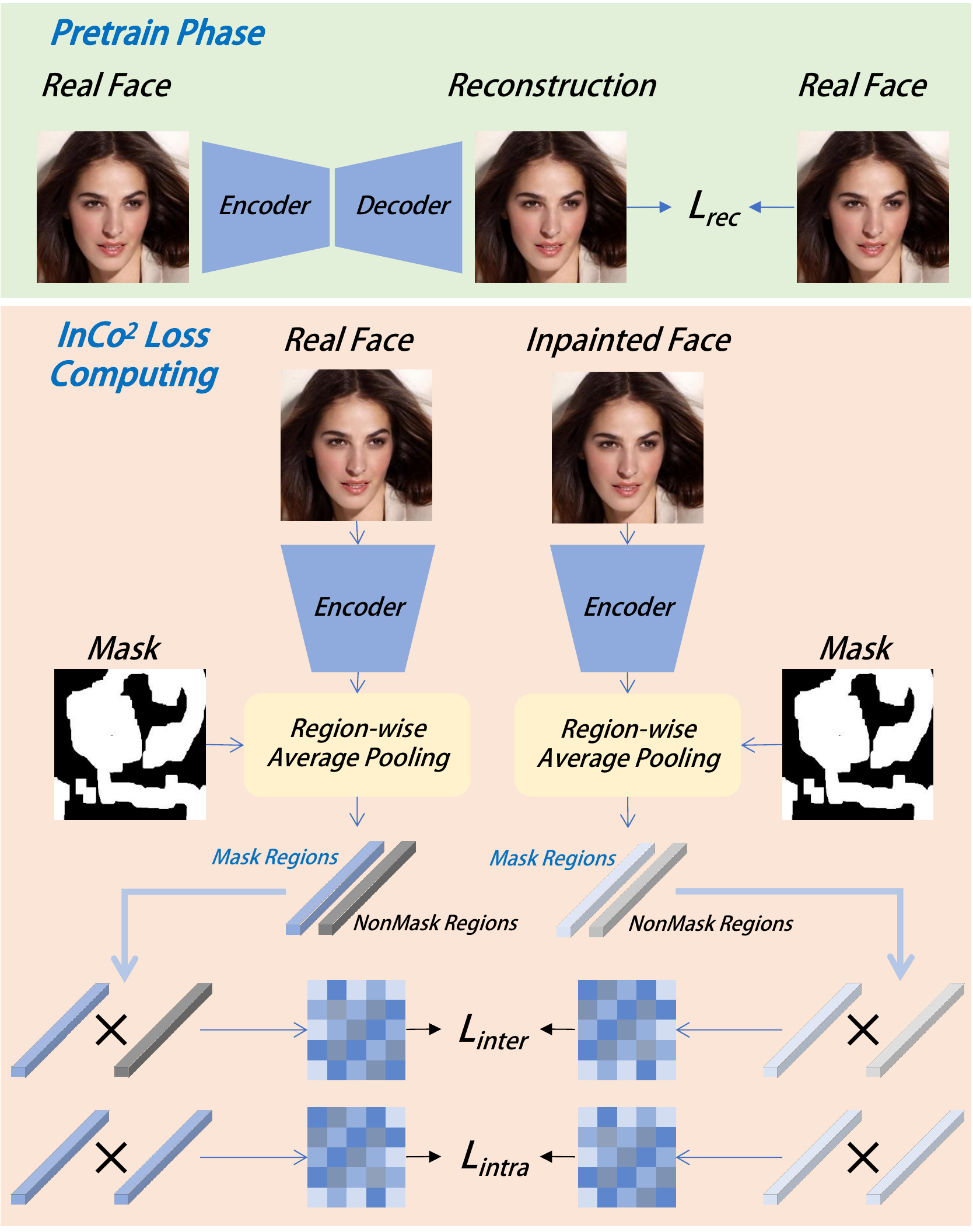}
  \caption{Detailed illustration of InCo$^2$ loss. In the pretrain phase, we train an encoder-decoder based reconstruction network. 
  In the Loss Computing, given a real or an inpainted face, we could use the pretrained encoder and the mask map to obtain two feature representations of the mask and non-mask regions by region-wise average pooling.
  Then we could obtain the InCo$^2$ Loss through similarity calculation.
  }
  \Description{InCo$^2$ Loss}
  \label{figure4}
\end{figure}

In this section, we present the objective functions of our method.
We first introduce the proposed InCo$^2$ loss in detail. 
Then we describe the semantic loss \cite{li2017generative} appropriately applied in our method. 
Finally, we illustrate the reconstruction loss, perceptual loss, style loss, adversarial loss and total variation loss, inherited from the previous generation methods\cite{liu2018image, liu2021pd, peng2021generating, wan2021high}.

\textbf{InCo$^2$ Loss.}  
In face inpainting, it is reasonable to focus on corrupted region reconstruction.
Meanwhile, coordinated face inpainting between the completed and the non-corrupted areas is also a key point.
In this paper, we propose InCo$^2$ Loss to further constrain the feature consistency for face coordination.
Specifically, the \textbf{InCo$^2$ Loss}  includes a
pair of similarity based losses,
\textbf{In}tra-class \textbf{Co}ordination loss and \textbf{In}ter-class \textbf{Co}ordination loss. 
The intra-class coordination requires the coordinated relationships among the various semantic features in corrupted regions. 
Similarly, inter-class coordination requires coordinated feature relationships between the
completed regions and the non-corrupted regions.
In the face inpainting task, we believe that \textbf{the rationality among the face regions is more important than the maintenance of pixel-level face consistency}.
Therefore, we design the similarity matrix based constraints, which could represent the correlations within features and the correlations between different features. 

Concretely, there are two steps to establish the InCo$^2$ Loss, as shown in Figure \ref{figure4}.
In the pretrain phase, we employ an encoder-decoder based reconstruction network to obtain the implicit representations of human face from the middle layer. 
Then we employ the pretrained encoder and the mask to obtain two representations $\mathcal{M}(\cdot)$ and $\hat{\mathcal{M}}(\cdot)$ of the mask and non-mask regions by region-wise average pooling\cite{zhu2020sean}, respectively.
The intra-class coordination loss and the inter-class coordination loss are defined as:
\begin{equation}
\mathcal{L}_{intra}=\left\| \mathcal{M}(\hat{I})\times \mathcal{M}(\hat{I})^{T}-\mathcal{M}(I_{gt})\times \mathcal{M}(I_{gt})^{T} \right\|_{1}   \label{eq6}
\end{equation}
\begin{equation}
\mathcal{L}_{inter}=\left\| \mathcal{M}(\hat{I})\times \hat{\mathcal{M}}(\hat{I})^{T}-\mathcal{M}(I_{gt})\times \hat{\mathcal{M}}(I_{gt})^{T} \right\|_{1}   \label{eq7}
\end{equation}

Therefore, InCo$^2$ Loss is defined as:
\begin{equation}
\mathcal{L}_{InCo^{2}}= \mathcal{L}_{intra} +   \mathcal{L}_{inter} \label{eq8}
\end{equation}

\textbf{Semantic Loss.}
Since the semantic map $S$ obtained from coarse face $P$ may bring some errors, we employ semantic loss $\mathcal{L}_{sem}$ to refine their influences, which computes the Cross Entropy of parsing maps between the completed image $\hat{I}$ and ground truth $I_{gt}$.
\begin{equation}
\mathcal{L}_{sem}= \mathbb{CE}(\mathbb{P}(I_{gt}),  \mathbb{P}(\hat{I}))  \label{eq9}
\end{equation}
where $\mathbb{P}$ denotes the inference process of face parser.

\textbf{Reconstruction Loss.}
The reconstruction loss $\mathcal{L}_{rec}$ calculates the $L1$ distance between the completed image $\hat{I}$ and ground truth $I_{gt}$, which encourages the $\hat{I}$ to be similar with $I_{gt}$ at the pixel level.
\begin{equation}
\mathcal{L}_{rec}= \left\| \hat{I} - I_{gt} \right\|_{1} \label{eq10}
\end{equation}

\textbf{Perceptual Consistency Loss.}
The perceptual loss $\mathcal{L}_{perc}$ measures the $L1$ distance between $\hat{I}$ and $I_{gt}$ in the feature space, which penalizes the perceptual and semantic discrepancy.  
\begin{equation}
\mathcal{L}_{perc}=\sum_{i}^{}\left\| \phi _{i}(\hat{I})- \phi _{i}(I_{gt})\right\|_{1}  \label{eq11}
\end{equation}
where $\phi_{i}(\cdot)$ denotes the activation of the $i$th layer from the pre-trained VGG-19 network \cite{Simonyan15}. 

\textbf{Style Consistency Loss.}
The style loss $\mathcal{L}_{style}$ calculates the statistical errors between the features of $\hat{I}$ and $I_{gt}$ to constrain the style consistency. 
\begin{equation}
\mathcal{L}_{style}=\sum_{i}^{}\left\| \mathbb{G}(\phi _{i}(\hat{I})) - \mathbb{G}(\phi _{i}(I_{gt})) \right\|_{1} \label{eq12}
\end{equation}
where $\mathbb{G}$ denotes the Gram matrix.

\textbf{Adversarial Loss.}
We employ the discriminator $D$ in PatchGAN \cite{isola2017image} to match distributions between $\hat{I}$ and $I_{gt}$, which promotes the generator to generate realistic images.
\begin{equation}
\mathcal{L}_{adv}=\mathbb{E}_{I_{gt}}[log(D(I_{gt}))]+\mathbb{E}_{\hat{I}}[log(1-D(\hat{I}))] \label{eq13}
\end{equation}

\textbf{Total Variation Loss.}
We adopt a total variation loss $\mathcal{L}_{tv}$ to smooth the completed image $\hat{I}$.
\begin{equation}
\mathcal{L}_{tv}= \left\| \hat{I} \right\|_{tv} \label{eq14}
\end{equation}

In summary, the overall objective function can be formulated as:
\begin{equation}
\begin{aligned}
\mathcal{L}_{total} =& \lambda _{1}\mathcal{L}_{InCo^{2}} +  \lambda _{2}\mathcal{L}_{sem} + \lambda _{3}\mathcal{L}_{rec} + \lambda _{4}\mathcal{L}_{perc} 
\\& + \lambda _{5}\mathcal{L}_{style} + \lambda _{6}\mathcal{L}_{adv} + \lambda _{7}\mathcal{L}_{tv} 
\end{aligned}
\label{eq15}
\end{equation}
where $\lambda _{i,\left\{ i=1,2,...,7 \right\}}$ are hyper-parameters to balance each item.

\section{EXPERIMENTS}
In this section, we first introduce the experimental settings, which include datasets, evaluation metrics and implementation details. 
Then we illustrate and analyze our experimental results.

\subsection{Experimental Settings}

\textbf{Datasets and Evaluation Metrics.}
We evaluate the proposed method on CelebA-HQ \cite{karras2018progressive} and FFHQ \cite{karras2019style}. 
We follow the split in \cite{yu2019free} to produce 28,000 training images and 2,000 validation images in CelebA-HQ.
For FFHQ, we preserve the last 2,000 images for test, and use the rest images for train.
Irregular masks provided by \cite{liu2018image} are employed for both training and evaluation. 
The $L1$ error, Fréchet Inception Distance (FID) \cite{heusel2017gans}, Peak Signal-to-Noise Ratio (PSNR), Structure Similarity (SSIM) \cite{wang2004image} are used to evaluate the quality of the results. 
The $L1$ error, PSNR and SSIM compare the differences between the completed image and ground truth. 
The FID calculates the distance of feature distributions between the completed face and ground truth.

\textbf{Implementation Details.}
The proposed method is implemented in PyTorch with 4 Nvidia Titan Xp GPUs. 
The image and mask are resized to 256 $\times$ 256 for training and evaluation. 
Our model is optimized using Adam optimizer with $\beta_{1} $=0.9 and $\beta_{2} $=0.99. 
We train the model for 45 epochs with the batchsize of 8. 
The learning rate is set to 2e-4. 
For coarse network, We replace the vanilla convolution of the U-Net architecture \cite{ronneberger2015u} with partial convolution \cite{liu2018image} as the coarse network.
The coarse network is trained on CelebA-HQ for 100 epochs, and other settings are the same as the MDRNets.
For the reconstruction network in Figure \ref{figure4}, we train the network for 30 epochs, with the same settings as above.

\subsection{Qualitative Analysis}

\begin{figure*}[h]
  \centering
  \includegraphics[width=0.95\textwidth]{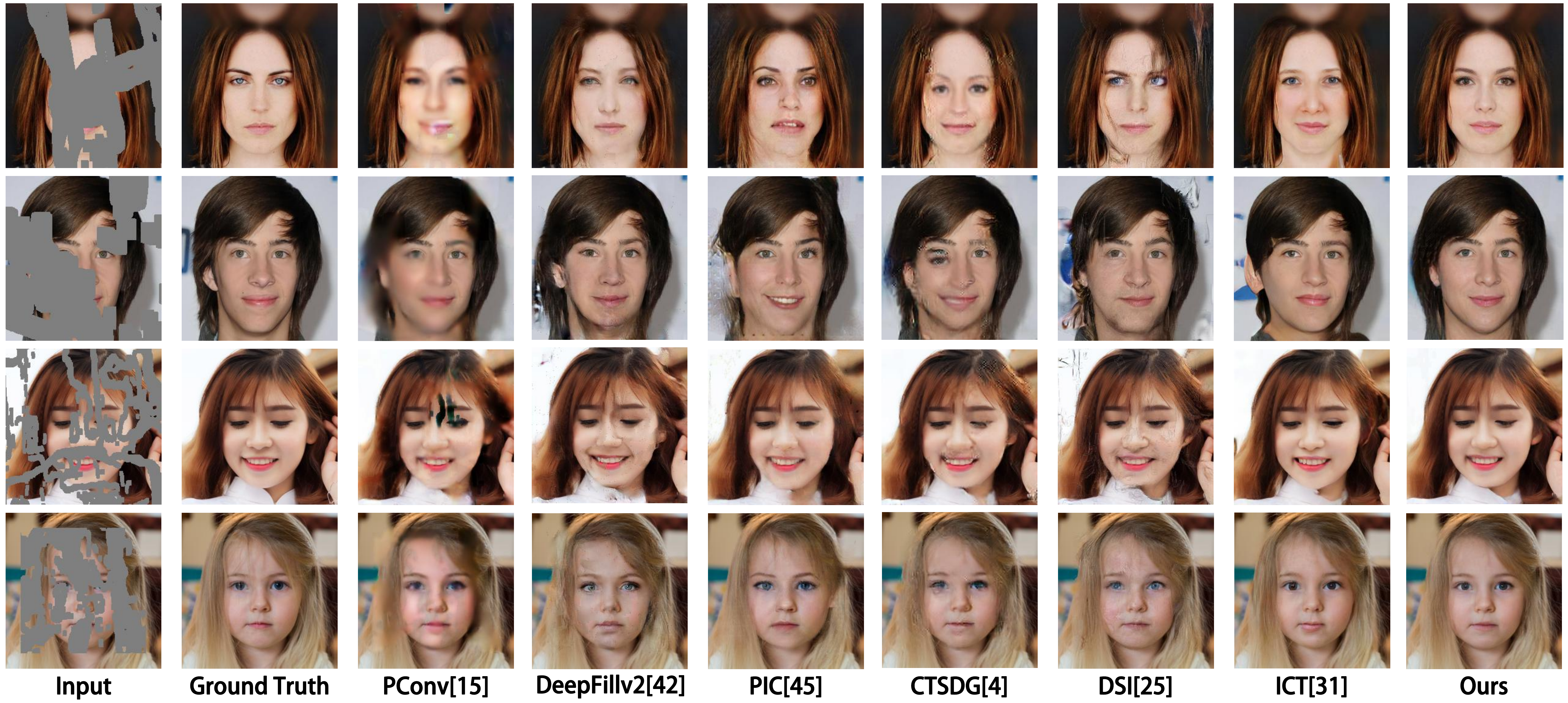}
  \vspace{-2mm}
  \caption{Randomly sampled results of our MDRNets compared with the previous SOTA face inpainting methods.}
  \Description{}
  \vspace{-2mm}
  \label{figure5}
\end{figure*}

We compare our methods with previous state-of-the-art approaches, including PConv \cite{liu2018image}, DeepFillv2 \cite{yu2019free}, PIC \cite{zheng2019pluralistic}, CTSDG \cite{guo2021image}, DSI \cite{peng2021generating} and ICT \cite{wan2021high}. 
All the results are obtained by using pre-trained models or implementation code published by the authors. 
We show the results of qualitative comparisons in Figure \ref{figure5}.
PConv and DeepFillv2 generate blurry results since these models can not capture valid contextual information.
PIC generates reasonable facial structures. 
However, the results of PIC suffer from artifacts due to the lack of adequate correlation.
CTSDG and DSI obtain distorted faces since these models use low-level structural information, which is incomplete in wide corrupted regions.
ICT could generate relatively satisfactory results. 
However, the results of ICT still have limitations in detailed textures since the model cannot perform fine restoration of each semantics.
Compared with these methods, our model achieves better results on both detailed textures and face coordination.
More qualitative results are presented in the supplementary materials.

\subsection{Quantitative Analysis}

\begin{table*}[h]
\centering
\vspace{-1mm}
\caption{Quantitative comparisons with SOTA methods on CelebA-HQ and FFHQ datasets. ($\downarrow$ Lower is better. $\uparrow$ Higher is better)}
\label{tab1} 
\resizebox{.98\textwidth}{!}{

\begin{tabular}{l|c|ccc|ccc|ccc|ccc}
\hline
\multicolumn{1}{c|}{\multirow{2}{*}{\textbf{Methods}}} & \multirow{2}{*}{\textbf{Dataset}}                         & \multicolumn{3}{c|}{$\textbf{$L$1}(\textbf{\%})\downarrow$} & \multicolumn{3}{c|}{$\textbf{FID}\downarrow$}     & \multicolumn{3}{c|}{$\textbf{PSNR} \uparrow$}       & \multicolumn{3}{c}{$\textbf{SSIM} \uparrow$}     \\ \cline{3-14} 
\multicolumn{1}{c|}{}                                  &                                                           & 1-20\%             & 20-40\%            & 40-60\%           & 1-20\%         & 20-40\%        & 40-60\%         & 1-20\%          & 20-40\%         & 40-60\%         & 1-20\%         & 20-40\%        & 40-60\%        \\ \hline \hline
PConv \cite{liu2018image}                            & \multirow{7}{*}{CelebA-HQ \cite{karras2018progressive}} & 1.131              & 2.311              & 4.363             & 12.716         & 27.957         & 42.594          & 32.240          & 26.085          & 21.900          & 0.941          & 0.862          & 0.762          \\
DeepFillv2 \cite{yu2019free}                         &                                                           & 0.788              & 2.066              & 3.968             & 9.766          & 22.793         & 29.243          & 32.700          & 25.998          & 21.943          & 0.944          & 0.848          & 0.736          \\
PIC \cite{zheng2019pluralistic}                      &                                                           & 0.780              & 2.036              & 4.311             & 4.190          & 11.035         & 21.360          & 33.006          & 25.961          & 21.263          & 0.951          & 0.859          & 0.730          \\
CTSDG \cite{guo2021image}                            &                                                           & 1.350              & 2.213              & 3.900             & 9.171          & 14.324         & 22.889          & 32.198          & 26.823          & 22.490          & 0.927          & 0.856          & 0.747          \\
DSI \cite{peng2021generating}                        &                                                           & 0.820              & 2.077              & 4.149             & 9.037          & 20.327         & 29.040          & 32.699          & 26.107          & 21.708          & 0.938          & 0.831          & 0.704          \\
ICT \cite{wan2021high}                               &                                                           & 0.949              & 2.004              & 3.901             & 3.136          & 8.715          & 16.747          & 33.416          & 26.639          & 22.013          & 0.959          & 0.879          & 0.765          \\
Ours                                                   &                                                           & \textbf{0.585}     & \textbf{1.451}     & \textbf{2.937}    & \textbf{2.369} & \textbf{6.410} & \textbf{12.086} & \textbf{35.772} & \textbf{28.669} & \textbf{24.177} & \textbf{0.968} & \textbf{0.900} & \textbf{0.800} \\ \hline \hline
PConv \cite{liu2018image}                            & \multirow{7}{*}{FFHQ \cite{karras2019style}}            & 0.720              & 2.178              & 4.411             & 12.208         & 30.403         & 45.709          & 32.592          & 25.422          & 21.237          & 0.955          & 0.867          & 0.761          \\
DeepFillv2 \cite{yu2019free}                         &                                                           & 0.715              & 2.104              & 4.250             & 12.062         & 29.276         & 40.295          & 32.428          & 25.470          & 21.301          & 0.946          & 0.845          & 0.725          \\
PIC \cite{zheng2019pluralistic}                      &                                                           & 0.709              & 2.099              & 4.573             & 5.411          & 14.344         & 27.334          & 32.640          & 25.490          & 20.819          & 0.952          & 0.854          & 0.719          \\
CTSDG \cite{guo2021image}                            &                                                           & \textbf{0.419}     & 1.532              & 3.569             & 3.916          & 13.477         & 28.495          & 34.946          & 27.044          & 22.272          & 0.968          & 0.888          & 0.765          \\
DSI \cite{peng2021generating}                        &                                                           & 0.746              & 2.067              & 4.340             & 10.483         & 25.772         & 39.127          & 32.659          & 25.780          & 21.241          & 0.941          & 0.834          & 0.702          \\
ICT \cite{wan2021high}                               &                                                           & 0.982              & 2.085              & 4.036             & 3.244          & 8.360          & 14.149          & 33.172          & 26.373          & 21.809          & 0.959          & 0.877          & 0.762          \\
Ours                                                   &                                                           & 0.470              & \textbf{1.395}     & \textbf{3.068}    & \textbf{2.473} & \textbf{7.170} & \textbf{13.748} & \textbf{36.046} & \textbf{28.333} & \textbf{23.575} & \textbf{0.972} & \textbf{0.903} & \textbf{0.797} \\ \hline
\end{tabular}
\vspace{-1mm}
}
\end{table*}

As shown in Table \ref{tab1}, we quantitatively evaluate the proposed method at irregular mask ratios of 1-20\%, 20-40\% and 40-60\%. 
As we can see, our proposed method outperforms other State-Of-The-Art methods on CelebA-HQ and FFHQ datasets. 
Especially, under the largest mask ratio, our method has \textbf{distinct improvements} compared with other methods.
Specifically, the $L1$ error and FID are reduced by 0.963\% and 4.661. 
Meanwhile, the PSNR and SSIM are improved by 1.687 and 0.035, compared to the sub-optimal result on the CelebA-HQ.
Similarly, the $L1$ error and FID are reduced by 0.501\% and 0.401. 
The PSNR and SSIM are improved by 1.303 and 0.032, compared to the sub-optimal result on the FFHQ.
The above results demonstrate the superiority of our method in coordinated face inpainting, especially with large masked regions.

\subsection{Ablation Study}

In this section, we perform extensive experiments to verify the effectiveness of each module and loss in our model. 
Then we conduct memory design ablation analysis. 
All the ablation experiments are performed on the CelebA-HQ dataset.

\begin{table}[]
\centering
\vspace{-1mm}
\caption{The evaluation results of Module Ablation.}
\label{tab2} 
\resizebox{1\linewidth}{!}{

\begin{tabular}{l|c|ccc}
\hline
\multicolumn{1}{c|}{\multirow{2}{*}{\textbf{Metrics}}} & \multirow{2}{*}{\textbf{\begin{tabular}[c]{@{}c@{}}Mask\\ Ratio\end{tabular}}} & \multicolumn{3}{c}{\textbf{Models}}                                                                                                     \\ \cline{3-5} 
\multicolumn{1}{c|}{}                                  &                                                                                & \multicolumn{1}{c|}{\begin{tabular}[c]{@{}c@{}}w/o MREN + w/o DMM\end{tabular}} & \multicolumn{1}{c|}{w/o MREM}       & Full            \\ \hline
$\textbf{$L$1}(\textbf{\%})\downarrow$                 & \multirow{4}{*}{1-20\%}                                                        & \multicolumn{1}{c|}{0.625}                                                      & \multicolumn{1}{c|}{0.595}          & \textbf{0.585}  \\
$\textbf{FID}\downarrow$                               &                                                                                & \multicolumn{1}{c|}{2.698}                                                      & \multicolumn{1}{c|}{\textbf{2.221}} & 2.369           \\
$\textbf{PSNR} \uparrow$                               &                                                                                & \multicolumn{1}{c|}{35.118}                                                     & \multicolumn{1}{c|}{35.467}         & \textbf{35.772} \\
$\textbf{SSIM} \uparrow$                               &                                                                                & \multicolumn{1}{c|}{0.963}                                                      & \multicolumn{1}{c|}{0.967}          & \textbf{0.968}  \\ \hline
$\textbf{$L$1}(\textbf{\%})\downarrow$                 & \multirow{4}{*}{20-40\%}                                                       & \multicolumn{1}{c|}{1.550}                                                      & \multicolumn{1}{c|}{1.473}          & \textbf{1.451}  \\
$\textbf{FID}\downarrow$                               &                                                                                & \multicolumn{1}{c|}{7.205}                                                      & \multicolumn{1}{c|}{6.529}          & \textbf{6.410}  \\
$\textbf{PSNR} \uparrow$                               &                                                                                & \multicolumn{1}{c|}{28.283}                                                     & \multicolumn{1}{c|}{28.470}         & \textbf{28.669} \\
$\textbf{SSIM} \uparrow$                               &                                                                                & \multicolumn{1}{c|}{0.890}                                                      & \multicolumn{1}{c|}{0.897}          & \textbf{0.900}  \\ \hline
$\textbf{$L$1}(\textbf{\%})\downarrow$                 & \multirow{4}{*}{40-60\%}                                                       & \multicolumn{1}{c|}{3.063}                                                      & \multicolumn{1}{c|}{2.959}          & \textbf{2.937}  \\
$\textbf{FID}\downarrow$                               &                                                                                & \multicolumn{1}{c|}{13.737}                                                     & \multicolumn{1}{c|}{13.359}         & \textbf{12.086} \\
$\textbf{PSNR} \uparrow$                               &                                                                                & \multicolumn{1}{c|}{24.017}                                                     & \multicolumn{1}{c|}{24.080}         & \textbf{24.177} \\
$\textbf{SSIM} \uparrow$                               &                                                                                & \multicolumn{1}{c|}{0.789}                                                      & \multicolumn{1}{c|}{0.798}          & \textbf{0.800}  \\ \hline
\end{tabular}

}
\end{table}

\begin{figure}[h]
  \centering
  \vspace{-1mm}
  \includegraphics[width=0.80\linewidth]{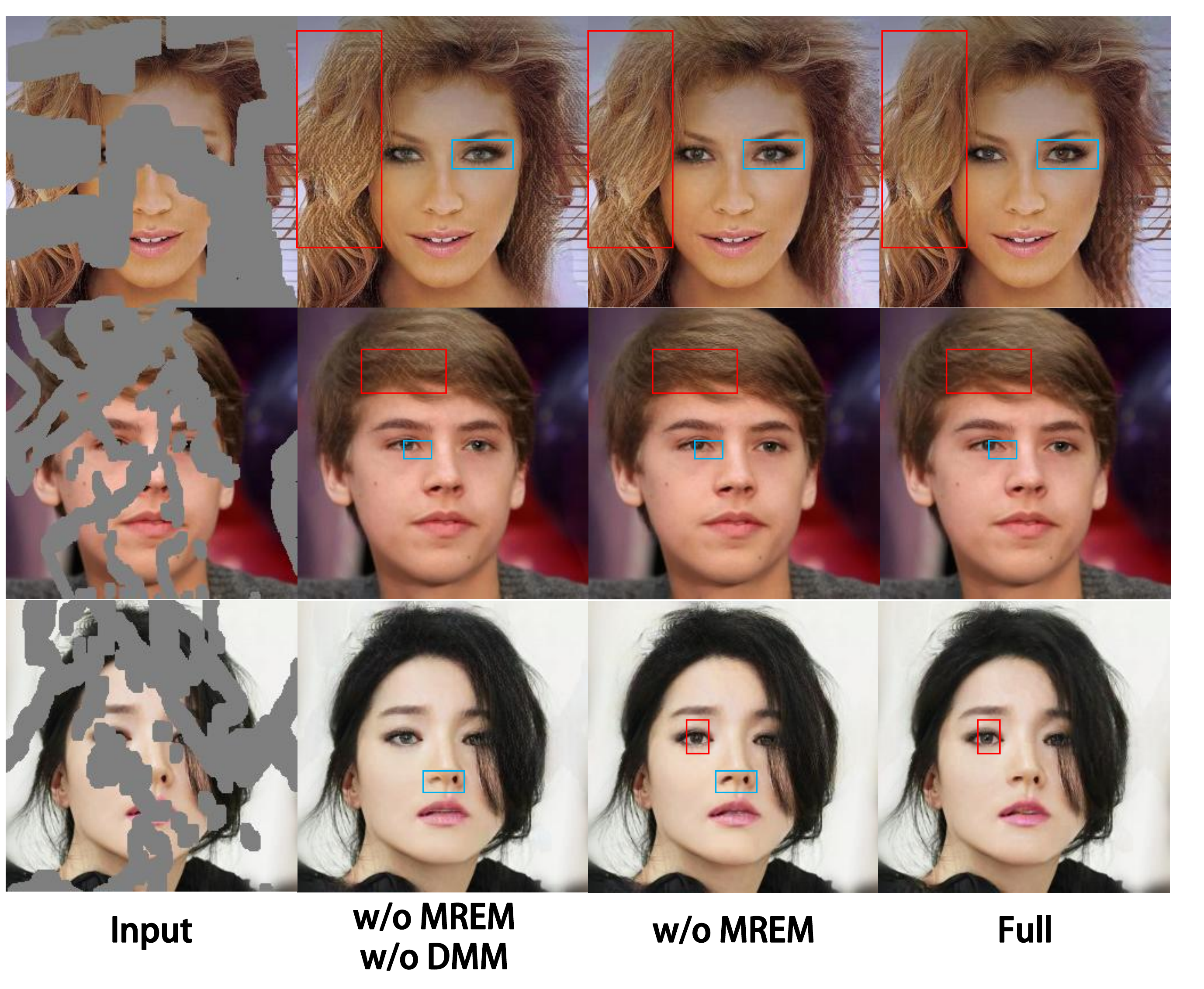}
  \vspace{-3mm}
  \caption{The qualitative comparisons of module ablation. Zoom in for better details.}
  \vspace{-3mm}
  \Description{.}
  \label{figure6}
\end{figure}

\textbf{Module Ablation.}
We further perform module ablation to demonstrate the effectiveness of each module. 
There are three models with different settings for experimental comparison: 
1). \textbf{w/o MREN + w/o DMM.} This model removes the MREN and DMM. The features extracted by the style encoder are injected into the Guided Refinement Network directly.
2). \textbf{w/o MREN.} This model removes the MREN and uses the fusion features after DMM to inject.  
3). \textbf{Full.} Our proposed modules are all used in experiments. 
The module ablation results are shown in Table \ref{tab2}. 
Figure \ref{figure6} also shows qualitative comparisons of module ablation.
The \textbf{w/o MREN + w/o DMM.} model is difficult to recover detailed textures at the semantic level, especially when some semantic categories are completely masked (e.g. the eyes of the first person in Figure \ref{figure6} are more blurred than the \textbf{Full} model. ) 
Meanwhile, the \textbf{w/o MREN} recovers some detailed textures, but the faces suffer from coordination issues.
In addition, the \textbf{Full} model achieved satisfactory results both in detailed textures and coordination. 
Finally, the \textbf{Full} model achieves the best performance. 
The above experimental results demonstrate that all our proposed modules are effective. 
\begin{table}[]
\centering
\vspace{-1mm}
\caption{The evaluation results of Loss Ablation.}
\vspace{-2mm}
\label{tab3} 
\resizebox{.98\linewidth}{!}{

\begin{tabular}{l|c|cccc}
\hline
\multicolumn{1}{c|}{\multirow{2}{*}{\textbf{Metrics}}} & \multirow{2}{*}{\textbf{\begin{tabular}[c]{@{}c@{}}Mask\\ Ratio\end{tabular}}} & \multicolumn{4}{c}{\textbf{Models}}                                                                                                                              \\ \cline{3-6} 
\multicolumn{1}{c|}{}                                  &                                                                                & \multicolumn{1}{c|}{w/o $\mathcal{L}_{sem}$} & \multicolumn{1}{c|}{w/o $\mathcal{L}_{intra}$} & \multicolumn{1}{c|}{w/o $\mathcal{L}_{inter}$} & Full            \\ \hline
$\textbf{$L$1}(\textbf{\%})\downarrow$                 & \multirow{4}{*}{1-20\%}                                                        & \multicolumn{1}{c|}{0.589}                   & \multicolumn{1}{c|}{0.587}                     & \multicolumn{1}{c|}{0.588}                     & \textbf{0.585}  \\
$\textbf{FID}\downarrow$                               &                                                                                & \multicolumn{1}{c|}{2.400}                   & \multicolumn{1}{c|}{2.472}                     & \multicolumn{1}{c|}{2.421}                     & \textbf{2.369}  \\
$\textbf{PSNR} \uparrow$                               &                                                                                & \multicolumn{1}{c|}{35.676}                  & \multicolumn{1}{c|}{35.646}                    & \multicolumn{1}{c|}{35.639}                    & \textbf{35.772} \\
$\textbf{SSIM} \uparrow$                               &                                                                                & \multicolumn{1}{c|}{0.967}                   & \multicolumn{1}{c|}{0.967}                     & \multicolumn{1}{c|}{0.967}                     & \textbf{0.968}  \\ \hline
$\textbf{$L$1}(\textbf{\%})\downarrow$                 & \multirow{4}{*}{20-40\%}                                                       & \multicolumn{1}{c|}{1.461}                   & \multicolumn{1}{c|}{1.452}                     & \multicolumn{1}{c|}{1.462}                     & \textbf{1.451}  \\
$\textbf{FID}\downarrow$                               &                                                                                & \multicolumn{1}{c|}{6.541}                   & \multicolumn{1}{c|}{7.011}                     & \multicolumn{1}{c|}{6.874}                     & \textbf{6.410}  \\
$\textbf{PSNR} \uparrow$                               &                                                                                & \multicolumn{1}{c|}{28.594}                  & \multicolumn{1}{c|}{28.631}                    & \multicolumn{1}{c|}{28.528}                    & \textbf{28.669} \\
$\textbf{SSIM} \uparrow$                               &                                                                                & \multicolumn{1}{c|}{0.898}                   & \multicolumn{1}{c|}{0.897}                     & \multicolumn{1}{c|}{0.897}                     & \textbf{0.900}  \\ \hline
$\textbf{$L$1}(\textbf{\%})\downarrow$                 & \multirow{4}{*}{40-60\%}                                                       & \multicolumn{1}{c|}{2.954}                   & \multicolumn{1}{c|}{2.939}                     & \multicolumn{1}{c|}{2.961}                     & \textbf{2.937}  \\
$\textbf{FID}\downarrow$                               &                                                                                & \multicolumn{1}{c|}{12.367}                  & \multicolumn{1}{c|}{13.721}                    & \multicolumn{1}{c|}{13.292}                    & \textbf{12.086} \\
$\textbf{PSNR} \uparrow$                               &                                                                                & \multicolumn{1}{c|}{24.111}                  & \multicolumn{1}{c|}{24.171}                    & \multicolumn{1}{c|}{24.171}                    & \textbf{24.177} \\
$\textbf{SSIM} \uparrow$                               &                                                                                & \multicolumn{1}{c|}{0.796}                   & \multicolumn{1}{c|}{0.795}                     & \multicolumn{1}{c|}{0.794}                     & \textbf{0.800}  \\ \hline
\end{tabular}
\vspace{-1mm}
}
\end{table}

\begin{figure}[h]
  \centering
  \vspace{-1mm}
  \includegraphics[width=1\linewidth]{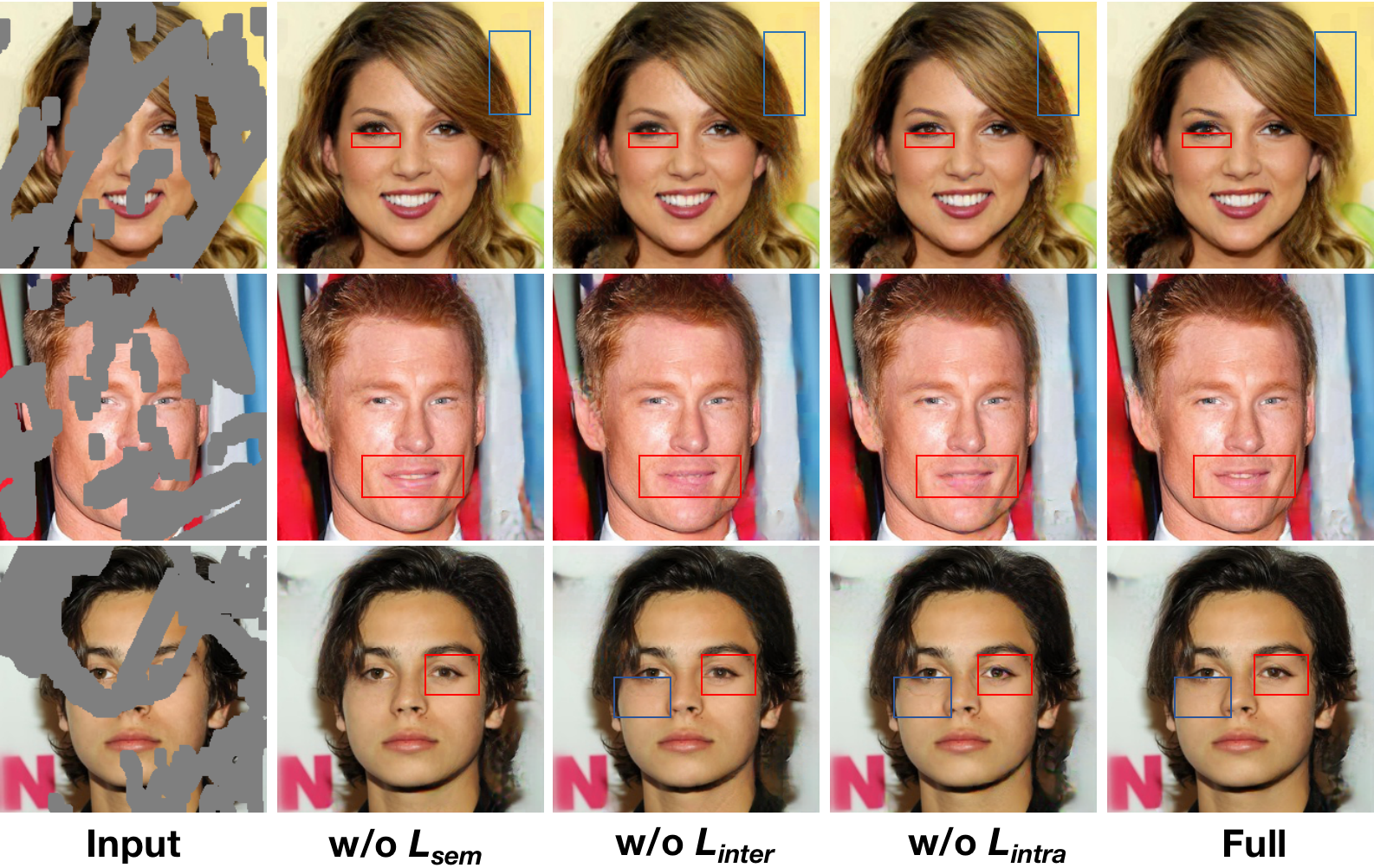}
  \vspace{-4mm}
  \caption{The qualitative comparisons of loss ablation. Zoom in for better details.}
  \vspace{-1mm}
  \Description{.}
  \label{figure7}
\end{figure}

\textbf{Loss Ablation.}
We conduct the loss ablation experiments to demonstrate the effectiveness of $\mathcal{L}_{sem}$ loss and the proposed $\mathcal{L}_{InCo^{2}}$ loss. 
The quantitative results of loss ablation are shown in Table \ref{tab3}.  
The \textbf{Full} model achieves the best performance on all metrics.
Meanwhile, the removal of any loss function will degrade the performance of the model integrally.
Figure \ref{figure7} shows the qualitative results of loss ablation. 
The w/o $\mathcal{L}_{sem}$ can lead to unclear semantic boundaries. 
Meanwhile, w/o $\mathcal{L}_{inter}$ can not maintain coordination between corrupted regions and non-corrupted regions.
Furthermore, w/o $\mathcal{L}_{intra}$ causes inconsistency within the corrupted regions.
The Full model could generate reasonable results.

\begin{table}[]
\centering
\caption{The evaluation results of Memory Design Ablation. $m$ denotes the slot number for each memory block. Non-Disentangled denotes using non-disentangled memory.}
\label{tab4} 
\resizebox{1\linewidth}{!}{

\begin{tabular}{l|c|ccccc|c}
\hline
\multicolumn{1}{c|}{\multirow{2}{*}{\textbf{Metrics}}} & \multirow{2}{*}{\textbf{\begin{tabular}[c]{@{}c@{}}Mask\\ Ratio\end{tabular}}} & \multicolumn{5}{c|}{\textbf{Disentangled}}                                                                                                                & \textbf{Non-Disentangled} \\ \cline{3-8} 
\multicolumn{1}{c|}{}                                  &                                                                                & \multicolumn{1}{c|}{$m$=32} & \multicolumn{1}{c|}{$m$=64}         & \multicolumn{1}{c|}{$m$=128}         & \multicolumn{1}{c|}{$m$=256} & $m$=512         & same volume as $m$=128    \\ \hline
$\textbf{$L$1}(\textbf{\%})\downarrow$                 & \multirow{4}{*}{1-20\%}                                                        & \multicolumn{1}{c|}{0.627}  & \multicolumn{1}{c|}{\textbf{0.584}} & \multicolumn{1}{c|}{0.585}           & \multicolumn{1}{c|}{0.600}   & 0.618           & 0.662                     \\
$\textbf{FID}\downarrow$                               &                                                                                & \multicolumn{1}{c|}{3.194}  & \multicolumn{1}{c|}{\textbf{2.308}} & \multicolumn{1}{c|}{2.369}           & \multicolumn{1}{c|}{2.670}   & 2.927           & 5.244                     \\
$\textbf{PSNR} \uparrow$                               &                                                                                & \multicolumn{1}{c|}{35.214} & \multicolumn{1}{c|}{35.718}         & \multicolumn{1}{c|}{\textbf{35.772}} & \multicolumn{1}{c|}{35.441}  & 35.397          & 34.987                    \\
$\textbf{SSIM} \uparrow$                               &                                                                                & \multicolumn{1}{c|}{0.963}  & \multicolumn{1}{c|}{0.968}          & \multicolumn{1}{c|}{\textbf{0.968}}  & \multicolumn{1}{c|}{0.966}   & 0.965           & 0.962                     \\ \hline
$\textbf{$L$1}(\textbf{\%})\downarrow$                 & \multirow{4}{*}{20-40\%}                                                       & \multicolumn{1}{c|}{1.547}  & \multicolumn{1}{c|}{1.452}          & \multicolumn{1}{c|}{\textbf{1.451}}  & \multicolumn{1}{c|}{1.494}   & 1.532           & 1.639                     \\
$\textbf{FID}\downarrow$                               &                                                                                & \multicolumn{1}{c|}{8.017}  & \multicolumn{1}{c|}{6.421}          & \multicolumn{1}{c|}{\textbf{6.410}}  & \multicolumn{1}{c|}{7.741}   & 7.626           & 15.452                    \\
$\textbf{PSNR} \uparrow$                               &                                                                                & \multicolumn{1}{c|}{28.344} & \multicolumn{1}{c|}{28.593}         & \multicolumn{1}{c|}{\textbf{28.669}} & \multicolumn{1}{c|}{28.391}  & 28.452          & 28.339                    \\
$\textbf{SSIM} \uparrow$                               &                                                                                & \multicolumn{1}{c|}{0.891}  & \multicolumn{1}{c|}{0.898}          & \multicolumn{1}{c|}{\textbf{0.900}}  & \multicolumn{1}{c|}{0.893}   & 0.893           & 0.893                     \\ \hline
$\textbf{$L$1}(\textbf{\%})\downarrow$                 & \multirow{4}{*}{40-60\%}                                                       & \multicolumn{1}{c|}{3.046}  & \multicolumn{1}{c|}{2.951}          & \multicolumn{1}{c|}{\textbf{2.937}}  & \multicolumn{1}{c|}{3.000}   & 3.037           & 3.177                     \\
$\textbf{FID}\downarrow$                               &                                                                                & \multicolumn{1}{c|}{14.124} & \multicolumn{1}{c|}{12.523}         & \multicolumn{1}{c|}{\textbf{12.086}} & \multicolumn{1}{c|}{15.557}  & 13.712          & 24.495                    \\
$\textbf{PSNR} \uparrow$                               &                                                                                & \multicolumn{1}{c|}{24.017} & \multicolumn{1}{c|}{24.079}         & \multicolumn{1}{c|}{24.177}          & \multicolumn{1}{c|}{23.996}  & \textbf{24.901} & 24.152                    \\
$\textbf{SSIM} \uparrow$                               &                                                                                & \multicolumn{1}{c|}{0.791}  & \multicolumn{1}{c|}{0.796}          & \multicolumn{1}{c|}{0.800}           & \multicolumn{1}{c|}{0.791}   & 0.793           & 0.800                     \\ \hline
\end{tabular}
\vspace{-1mm}
}
\end{table}

\textbf{Memory Design Ablation.}
The number of slots in each memory block is a question worth considering.
That is, how many slots do we need to store the latent vectors of each semantic .
In the Table \ref{tab4}, we present the results for different $m$. We can clearly see that the $m=128$ could obtain the best results in the disentangled memory. 
Meanwhile, we conduct a comparison on whether the memory is disentangled. 
According to the results, \textbf{the disentangled memory with $m=128$} has better performances in face inpainting.

\section{CONCLUSIONS}
In this paper, we propose Memory-Disentangled Refinement Networks (MDRNets) for coordinated face inpainting.
We propose two collaborative modules, the DMM to establish a group of disentangled memory and the MREM to enhance feature correlation.
Meanwhile, we design InCo$^2$ Loss, a pair of similarity based losses to better improve the inter-coordination between the corrupted and non-corrupted regions and enhance the intra-coordination in corrupted regions.
Extensive experiments conducted on CelebA-HQ and FFHQ datasets demonstrate the superiority of our MDRNets.

\clearpage

\bibliographystyle{ACM-Reference-Format}
\bibliography{ShowFace_1}

\appendix

\end{document}